\documentclass[conference]{IEEEtran}

\usepackage[numbers]{natbib}
\usepackage[nocompress]{cite}

\usepackage{microtype}
\usepackage{graphicx}
\usepackage{subcaption}
\usepackage{booktabs} 
\usepackage{multicol}

\usepackage{amsmath,amsfonts,bm}

\def\eqref#1{equation~\ref{#1}}

\def\1{\bm{1}}

\DeclareMathAlphabet{\mathsfit}{\encodingdefault}{\sfdefault}{m}{sl}
\SetMathAlphabet{\mathsfit}{bold}{\encodingdefault}{\sfdefault}{bx}{n}

\newcommand{\E}{\mathbb{E}}

\newcommand{\R}{\mathbb{R}}

\PassOptionsToPackage{hyphens}{url}\usepackage[hidelinks]{hyperref}

\usepackage{xspace}
\usepackage{xfrac}
\usepackage{caption}
\usepackage[dvipsnames]{xcolor}
\usepackage{float}
\usepackage{xcolor} 
\usepackage{tcolorbox}

\usepackage{amsmath}
\usepackage{amssymb}
\usepackage{mathtools}
\usepackage{amsthm}
\usepackage{bbm}
\usepackage{multirow}
\usepackage{enumitem}
\usepackage{tikz}
\usepackage{nicefrac}
\usetikzlibrary{shapes}
\usetikzlibrary{shapes.misc}
\tikzset{cross/.style={cross out, draw=black, minimum size=2*(#1-\pgflinewidth), inner sep=0pt, outer sep=0pt},
cross/.default={1pt}}
\usepackage{standalone}

\def\minop{\mathop{\rm min}\limits}
\def\maxop{\mathop{\rm max}\limits}
\def\argmaxop{\mathop{\rm arg\,max}\limits}

\newcommand{\norm}[1]{\left\|#1\right\|}

\let\vec\relax
\newcommand{\vec}[1]{\boldsymbol{#1}}

\makeatletter
\def\@IEEEsectpunct{.\ \,}
\def\paragraph{\@startsection{paragraph}{4}{\z@}{1.5ex plus 1.5ex minus 0.5ex}%
{0ex}{\normalfont\normalsize\bfseries}}
\makeatother

\usepackage{wrapfig}

\usepackage[capitalize,noabbrev]{cleveref}

\makeatletter
\newcommand{\linebreakand}{%
  \end{@IEEEauthorhalign}
  \hfill\mbox{}\par
  \mbox{}\hfill\begin{@IEEEauthorhalign}
}
\makeatother

\title{Scaling Compute Is Not All You Need\\for Adversarial Robustness}

\author{ \centering
\IEEEauthorblockN{Edoardo Debenedetti}
\IEEEauthorblockA{ETH Zurich}
\and
\IEEEauthorblockN{Zishen Wan}
\IEEEauthorblockA{Georgia Institute of Technology}
\and
\IEEEauthorblockN{Maksym Andriushchenko}
\IEEEauthorblockA{EPFL}
\and
\IEEEauthorblockN{Vikash Sehwag}
\IEEEauthorblockA{Princeton University}
\linebreakand
\IEEEauthorblockN{Kshitij Bhardwaj}
\IEEEauthorblockA{Lawrence Livermore National Laboratory}
\and 
\IEEEauthorblockN{Bhavya Kailkhura}
\IEEEauthorblockA{Lawrence Livermore National Laboratory}
}

\begin{document}

\maketitle

\begin{abstract}
    The last six years have witnessed significant progress in adversarially robust deep learning. As evidenced by the CIFAR-10 dataset category in RobustBench benchmark, the accuracy under $\ell_\infty$ adversarial perturbations improved from 44\% in \citet{Madry2018Towards} to 71\% in \citet{peng2023robust}. Although impressive, existing state-of-the-art is still far from satisfactory. It is further observed that best-performing models are often very large models adversarially trained by industrial labs with significant computational budgets. In this paper, we aim to understand: ``how much longer can computing power drive adversarial robustness advances?" To answer this question, we derive \emph{scaling laws for adversarial robustness} which can be extrapolated in the future to provide an estimate of how much cost we would need to pay to reach a desired level of robustness. We show that increasing the FLOPs needed for adversarial training does not bring as much advantage as it does for standard training in terms of performance improvements. Moreover, we find that some of the top-performing techniques are difficult to exactly reproduce, suggesting that they are not robust enough for minor changes in the training setup. Our analysis also uncovers potentially worthwhile directions to pursue in future research.
    Finally, we make our benchmarking framework (built on top of \texttt{timm}~\citep{rw2019timm}) publicly available to facilitate future analysis in efficient robust deep learning.

\end{abstract}

\section{Introduction}
Rapid advances in deep learning continue to improve performance on benchmarks across data modalities, achieving near-human, and in some cases even better than human performance. However, the challenge of achieving very high adversarial robustness remains elusive, despite substantial investmentment in robust learning mechanisms. While scaling of computational resources has become a de facto approach to drastically improve the performance of deep neural networks across numerous tasks~\citep{bahri2021explaining,bansal2022data,gordon2021data,kaplan2020scaling,krishnan2022quarl,sharma2022scaling,zhai2022scaling}, similar studies are nearly missing in the context of adversarial robustness. In this work, we take some first steps toward answering the question:
\begin{center}
    \emph{Is scaling \underline{compute} a viable solution\\for achieving adversarial robustness?}
\end{center}
Note that scaling of computational resources in deep learning has a direct increase in environmental impact. Thus we are further interested in a two-fold impact of scaling, i.e., whether continuing a ``scaling compute" strategy would provide proportional improvement in adversarial robustness or would it have a disproportional environmental impact without much benefit to adversarial robustness.

To achieve this objective, we carry out a systematic empirical exploration by adversarially training a large number of models on CIFAR-10 dataset, the most commonly studied dataset on this task. Specifically, we vary several factors such as model size, adversarial loss, attack steps, use of synthetic data, etc. Next, we analyze the adversarial robustness of the resulting adversarially trained model zoo. Contrary to the intuitive notion that greater computational power (i.e., deeper and more complex models) should inherently improve a model's ability to withstand adversarial attacks, our research reveals an interesting phenomenon: \textit{scaling up model sizes does not yield proportionate improvements in adversarial robustness}---setting adversarial training apart from standard training. This implies that exponentially more computational resources (and carbon emission) are required to achieve similar gains, making compute scaling an inefficient (if not infeasible) approach to solving this open problem. 

We argue that the adversarial robustness community must think beyond the traditional paradigm of simply increasing compute. We must explore innovative avenues for addressing this important challenge, one that has clearly defied conventional approaches and has profound implications for the security and reliability of modern machine learning systems. In this work, we highlight the limitations of the current strategies and advocate for more creative and holistic approaches to achieving adversarial robustness, urging the field to break new ground in pursuit of robust and resilient AI systems.

\section{Background and Related Work}

\subsection{Adversarial Robustness}\label{subsec:adv_robustness}
\paragraph{Background and Threat Model}
For a model $f: \mathcal{X} \rightarrow \R^C$, input $\vec{x} \in \mathcal{X}$, and label $y \in \{1, \ldots, C\}$, we say that an adversarial perturbation $\vec{\delta} \in \Delta$ is \textit{successful} if 
\begin{align}\label{eq:misclassification_condition}
    \argmaxop_{c \in \{1, \ldots, C\}} f(\vec{x}+\vec{\delta})_c \neq y. 
\end{align}
Following the previous established works \citep{croce2020robustbench, Madry2018Towards, szegedy2013intriguing}, we assume that the attacker has \textit{full knowledge} about $f$ (i.e., a white-box threat model), and we measure adversarial robustness via \textit{robust accuracy} for the $\ell_\infty$ norm, i.e., the fraction of points $x$ on which the model $f$ predicts the correct class $y$ for all possible perturbations $\delta$ from $\Delta = \{\vec{\delta}\in\mathcal{X}\ |\ \norm{\vec{\delta}}_\infty \leq \varepsilon\}$. 
We note that robustness to small $\ell_\infty$-perturbations is of interest in different contexts: for robustness to unseen perturbations \citep{kang2019transfer}, better gradient-based interpretability \citep{tsipras2018robustness}, feature learning \citep{Salman2020Do}, generalization performance \citep{xie2020adversarial}. 
Many current robust training approaches are derived from \textit{adversarial training} which is formulated as the following min-max robust optimization problem in \citet{Madry2018Towards}: 
\begin{align}
    \minop_{f} \E_{(\vec{x}, y) \sim D} \big[ \maxop_{\delta\in\Delta} \ell(f(\vec{x}+\vec{\delta}), y) \big],
    \label{eq:adv_training}
\end{align}
where the inner maximization is typically approximated by multiple iterations of projected gradient descent (PGD). Importantly, since PGD has to be performed on every iteration of training, this significantly increases the computational cost of robust training. 

\paragraph{Existing Robustness Benchmarks}
There are multiple popular robustness benchmarks and libraries in the community. However, importantly, none of these benchmarks take into account the computational costs needed to train state-of-the-art robust models. 
Most libraries focus primarily on robustness \textit{evaluation} and implement various white- and black-box adversarial attacks \citep{papernot2018cleverhans, foolbox, art2018, ding2019advertorch, melis2019secml, goodman2020advbox, li2020deeprobust}. 
Other libraries focus on implementing defenses, most prominently \citet{art2018} implement multiple defenses from the literature and \citet{Engstrom2019Robustness} provides an implementation of adversarial training. 
As for benchmarks, one of the first benchmarks for different \textit{attacks} is introduced in the works of \citet{Madry2018Towards} and \citet{Zhang2019Theoretically} whose models withstood many thorough robustness evaluations. 
RobustML (\url{https://www.robust-ml.org/}) was one of the early benchmarks that aimed at collecting robustness claims from the community for the most prominent defenses. 
\citet{croce2020robustbench} further introduced RobustBench, a popular adversarial robustness benchmark that includes a large repository of models and is based on AutoAttack \citep{croce2020reliable},
an ensemble of four hyperparameter-free attacks (white- and black-box) that has shown reliable performance over a large set of models from the literature. 
In our work, we follow \citet{croce2020robustbench} and also rely on AutoAttack for standardized measurement of $\ell_\infty$ adversarial robustness.

\subsection{Neural Scaling Laws}
\paragraph{Scaling in Deep Learning}
The exploration into neural scaling laws, which describe how the performance of neural networks changes with varying model size, data, and computational budgets, has been consistently observed and validated across diverse domains like image classification~\citep{bahri2021explaining,zhai2022scaling}, language modeling~\citep{kaplan2020scaling,sharma2022scaling}, and neural machine translation~\citep{bansal2022data,gordon2021data}. Recent works have found that scaling up the data size, the model size, and the training schedule often leads to substantially improved performance. 
\citet{kaplan2020scaling} elucidates the intricate relationship between model size, dataset size, and computational budget, spotlighting pivotal scaling laws foundational to the efficacy of expansive neural language models. This work also offers strategies to optimally allocate a predetermined compute budget.
In a bid to empirically gauge the advantages of scaling, \citep{alabdulmohsin2022revisiting} introduces a robust methodology to reliably deduce scaling law parameters from learning curves. This methodology, rooted in the \textit{extrapolation loss}, offers insights into how neural architecture variations influence scaling exponents. 
The work by~\citep{lohn2022ai} delves deeper into the pivotal role computation plays in AI evolution, presenting insights into potential inflection points where computational power might no longer yield substantial AI advancements.

\paragraph{Environmental Implications} 
The escalating computational requisites of deep learning invariably lead to surges in energy consumption.
\citet{strubell2019energy} quantifies the approximate financial and environmental costs associated with training language models and proposes recommendations to reduce costs and improve equity in NLP research and practice. 
\citet{thompson2020computational} offers an introspective look at the computational underpinnings of DL models, suggesting that advancements in DL models are increasingly tethered to computational growth which poses economic, technical, and environmental challenges that beckon more efficient computational strategies. 
\citet{schwartz2020green} emphasizes that the computational expenses of cutting-edge AI endeavors have grown by a factor of 300,000 in recent epochs, resulting in a very substantial carbon footprint. This underscores the imperative to prioritize efficiency as a cardinal metric, aligned with accuracy, to mitigate AI's environmental impact and bolster inclusivity.
Echoing these sentiments, \citet{henderson2020towards} champions the cause of carbon accountability in AI, introducing a transparent and standardized reporting paradigm. 
However, similar studies are nearly missing in the context of adversarial robustness. Therefore, in this work, we pioneer an exploration into the neural scaling laws governing adversarial robustness.

\section{Experimental Roadmap}
\label{sec:exp}
\subsection{Adversarial Training Setup}
\label{subsec:setup}

We run extensive experiments over several hyper-parameters, spanning design choices such as loss function, architecture, activation function, number of attack steps, and amount of data used. \textit{Overall, we train 320 models}, requiring more than ten thousand GPU hours of training.

\paragraph{Robust Losses} The first axis we ablate is the adversarial training loss: we compare the original Adversarial Training (AT) loss (introduced earlier in \Cref{subsec:adv_robustness}) to the TRADES~\citep{Zhang2019Theoretically} one. Briefly speaking, TRADES reformulates the original PGD loss as a regularization term to improve the trade-off between robust and standard accuracy. In order to do this, it has two loss components, weighted by a parameter $\beta$ used to balance the trade-off. The two components are the loss on the original input $\vec{x}$, and the loss on the adversarial example found by maximizing the Kullback–Leibler divergence between the model output given the original input $\vec{x}$ and given the adversarial example $\vec{x} + \vec{\delta}$.  Formally, the loss is:

\begin{align}
    \minop_{f} \E_{(\vec{x}, y) \sim D} \big[ \ell(f(\vec{x}), y) + \beta \maxop_{\delta\in\Delta} \ell_{KL}(f(\vec{x}), f(\vec{x}+\vec{\delta})) \big].
    \label{eq:trades}
\end{align}

Interestingly, the TRADES and the AT losses require different amounts of compute: for TRADES, we need to run one forward pass in addition to the number of forward and backward passes needed to compute the adversarial example for the given training input, while for the PGD loss, we need to run only as many forward passes as the number of attack iterations. However, we note that the additional forward pass required by TRADES gets amortized when the number $n$ of attack steps is larger. For example, for WideResNet-28-10 with $1$ attack steps, TRADES increases the number of FLOPs for one epoch by $1.3\times$, while it increases by only $1.08\times$ when using 10 iterations.

\paragraph{Models} We test several sizes of WideResNet~\citep{zagoruyko2016wide}, with different depths and widths, i.e., 28-10, 34-10, 34-20, and 70-16. This is a common architecture in the adversarial training literature, and these are the most used size configurations for this architecture~\citep{croce2020robustbench}. Studying these architecture sizes enables us to analyze models with a wide range of FLOPs per forward pass and parameters, from $10.5$ GFLOPs and $36M$ parameters for WideResNet-28-10 to $77.6$ GFLOPs and $266M$ parameters for WideResNet-70-16. Moreover, we also test two different activation functions: ReLU, used originally in WideResNet, and GELU~\citep{hendrycks2016gaussian}. Previous work observed that smooth activation functions improve robustness~\citep{xie2020smooth,debenedetti2023light}. However, also GELU comes at extra compute cost as it requires about $1.5\times$ the memory required by ReLU.

\paragraph{Attack Steps} We vary the number of attack iterations, training our models with $1$, $2$, $5$, $7$, and $10$ iterations. Increasing attack iterations leads to stronger attacks. Obviously, training models with more steps require more FLOPs as each attack iteration requires one additional forward and one additional backward pass over the model.

\paragraph{Number of Training Epochs and Exponential Moving Average} We follow both the original training set-up from \citet{Zhang2019Theoretically}, with 100 training epochs and no exponential moving average (EMA) of the model weights, and the set-up proposed by \citet{Gowal2020Uncovering}. The latter comprises training the model for 400 epochs with EMA of the model weights with a decay of $0.995$\footnote{Note that we do not test exponential moving average with 100 epochs because, due to the exponential mechanism, 100 epochs are often not enough to benefit from EMA.}. This setup, not only requires $4\times$ the number of FLOPs for the overall training, but it also requires $2\times$ the memory as we need to keep both the weights being trained and those being averaged through EMA.

\paragraph{Synthetic Data} Finally, we test the contribution achieved by synthetic data for adversarial training. Previous work shows that extra unlabeled data provides a significant boost in performance~\citep{Carmon2019Unlabeled} and that this holds also if the extra data is synthetically generated~\citep{gowal2021improving,sehwag2021robust,wang2023better}. For this reason, we also test how much synthetic data helps with robustness in all the aforementioned scenarios. We use 1M synthetic CIFAR-10 images released by~\citet{gowal2021improving}. However, we do not account for the synthetic data generation nor the generative model training in our FLOP count. This is mainly because there are many open-source pre-trained models that can be used for generating data, and, overall, the compute needed to generate the data is negligible compared to the overall training compute, and the synthetic data can be used across different training runs, hence the compute amortized.

\subsection{Metrics of Interest}

\subsubsection{Accuracy-Related Metrics} As we are interested in adversarial robustness, we report the AutoAttack accuracy of our models. AutoAttack~\citep{croce2020reliable}, an ensemble of four different attacks, is the de-facto standard to benchmark the robustness of deep learning vision models. However, as running AutoAttack is computationally expensive, during training we evaluate the models using PGD-40 on a subset of the test samples, and apply early-stopping based on this metric. This also mitigates potential overfitting of the overall results to AutoAttack.

\subsubsection{Efficiency-Related Metrics}

To quantify the computational efficiency and environmental cost of adversarial robustness, we perform an analysis of the FLOPs, electricity consumption, and carbon emissions required to train a variety of popular robust models. A detailed description of our methodology is as follows:

\paragraph{FLOPs} 
To gauge model training efficiency in terms of floating-point operations per second (FLOPs), we employ the DeepSpeed Flops Profiler\footnote{DeepSpeed Flops Profiler: https://www.deepspeed.ai/tutorials/flops-profiler/}. This measurement is acquired from all submodules of the model throughout the training process. We compute the average total number of FLOPs required for training the model (excluding the validation of the model performed after each epoch), hence accounting for number of epochs, model size, number of attack steps, etc. As DeepSpeed does not enable measuring the FLOPs required for the backward pass, we approximate those FLOPs to be $2\times$ the number of FLOPs needed for the forward pass~\citep{hobbhahn2021backpropagation}.

\paragraph{Electricity Consumption} 
During model training, we repeatedly query the NVIDIA System Management Interface to sample GPU power consumption and get the average over all samples. By averaging these values and integrating them with the training time and the number of GPUs utilized, we estimated the overall power consumption in kilowatt-hours (kWh). Additionally, we incorporated the Power Usage Effectiveness (PUE) coefficient - a metric that encapsulates the ancillary energy expended to sustain the computational infrastructure, primarily cooling. With a reference PUE coefficient of 1.58 (the global average for data centers in 2018), we subsequently computed the cumulative electricity expenditure for training a model, using the U.S. average electricity rate of \$0.12/kWh.

\paragraph{Carbon Emission} 
To encapsulate the carbon footprint, we adopted the metric of \emph{CO2-equivalents}. This metric facilitates a unified representation of the global warming potential intrinsic to diverse greenhouse gases, expressed as an equivalent quantity of $\text{CO}_2$~\citep{eggleston20062006}. We harnessed the Machine Learning Emissions Calculator~\citep{lacoste2019quantifying} for this purpose, integrating parameters like GPU power consumption (derived from the NVIDIA System Management Interface), model training time, chosen cloud provider, and the geographic locale of computation. For our evaluations, we selected the Google Cloud Platform as the cloud provider and the U.S. as the computational region (with an estimated carbon footprint of 566.3 gCO2/kWh), as shown in \Cref{sec:result}.

\begin{figure*}[t!]
    \centering
    \begin{subfigure}{0.3\textwidth}
        \includegraphics[width=\textwidth]{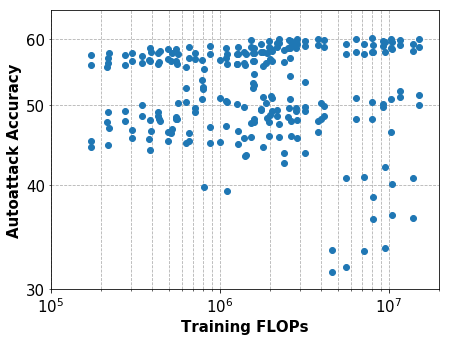}
        \caption{Training FLOPs vs Autoattack accuracy in log-log space.}
        \label{fig:algo1}
    \end{subfigure}
    \hspace{5pt}
    \begin{subfigure}{0.34\textwidth}
        \includegraphics[width=\textwidth]{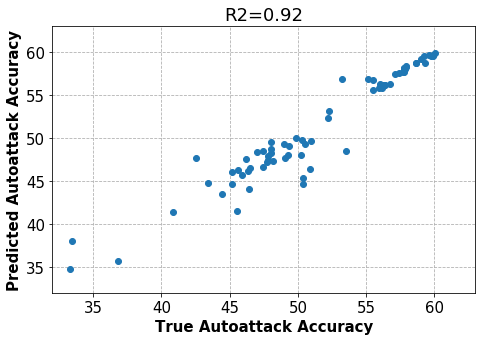}
        \caption{Performance of GBR regressor on autoattack accuracy prediction task.}
        \label{fig:algo2}
    \end{subfigure}
    \hspace{5pt}
    \begin{subfigure}{0.3\textwidth}
        \includegraphics[width=\textwidth]{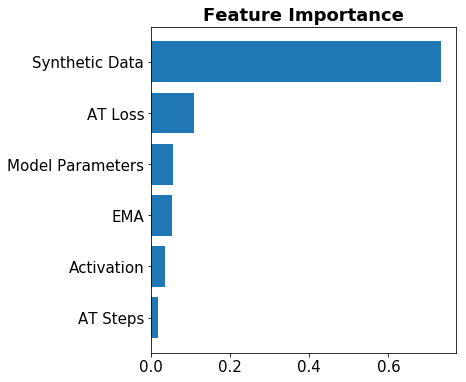}
        \caption{Feature importance of GBR model.}
        \label{fig:algo3}
    \end{subfigure}
    \caption{Analysis of various algorithmic changes on adversarial robustness.}
\end{figure*}

\section{Results}
\label{sec:result}

In this section,  we analyze the data generated in \Cref{sec:exp} to answer our questions of interest. As can be seen from \Cref{fig:algo1}, FLOP counts of our training runs span two orders of magnitude, and the generated data is sufficiently diverse.

\subsection{What Algorithmic Choices are Driving Adversarial Robustness Advances?}
\label{sec:algo}

In this section, we aim to unravel which algorithmic interventions are important for advancing adversarial robustness. We first try to predict how well a model (using a specific adversarial training recipe) can perform without actually training or testing it.
Specifically, we use number of model parameters, synthetic data $\in$ \{True, False\}, activation $\in$ \{ReLU, GELU\}, loss $\in$ \{PGD, TRADES\}, PGD steps $\in$ \{1, 2, 5, 7, 10\}, and exponential moving average (EMA) $\in$ \{True, False\} as input features to our predictive model. We use gradient boosting regression model with number of estimators equal to 50 and max depth equal to 5. We use a 70:30 train-test split in our experiment. 

In \Cref{fig:algo2}, we plot AutoAttack accuracy for unseen/holdout adversarial training configurations predicted by our model against ground truth values. We see that our model can predict the performance of unseen test configurations with high accuracy.  

\begin{tcolorbox}[colframe=black,colback=lightgray!30,boxrule=0.8pt,boxsep=4pt,left=1pt,right=1pt,top=0pt,bottom=0pt]
\noindent \emph{To the best of our knowledge, this is the first attempt to show that the adversarial robustness of a given training recipe can directly be predicted without going through a computationally expensive training-validation cycle.}
\end{tcolorbox}

This finding is expected to have broad implications for compute-efficient training~\citep{bartoldson2023compute} and reliable testing~\citep{zhang2020machine} of machine learning models.

Next, knowing which algorithmic knobs are important for predicting AutoAttack accuracy can be very useful in helping designers understand the problem and devising better approaches for improving the adversarial robustness. Thus, we analyze the feature importance of our predictive model in \Cref{fig:algo3}. Feature importance provides a score that indicates how useful (or valuable) each feature is for our predictive model. These importance scores are calculated individually for each feature in our dataset, enabling the features to be ranked and compared to each other in \Cref{fig:algo3}.

\begin{tcolorbox}[colframe=black,colback=lightgray!30,boxrule=0.8pt,boxsep=4pt,left=1pt,right=1pt,top=0pt,bottom=0pt]
\noindent \emph{We hypothesize that the use of synthetic data, training loss, and the number of model parameters are the most important driving forces behind the adversarial robustness advances. }
\end{tcolorbox}

\begin{figure}[h]
    \centering
    \includegraphics[width=0.5\textwidth]{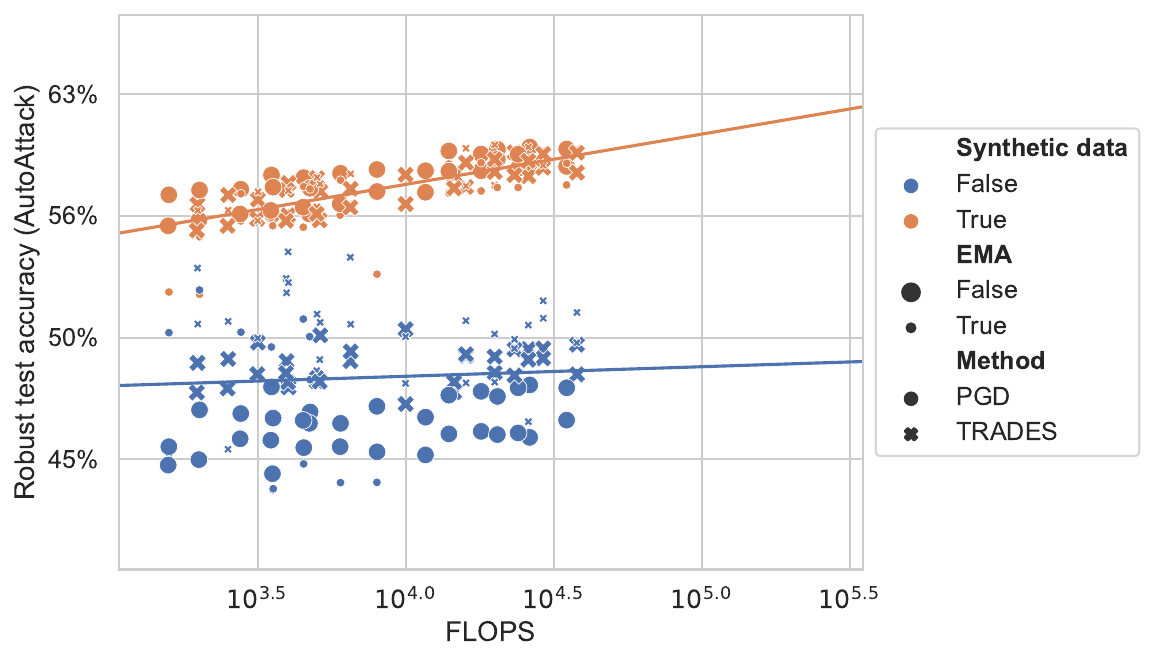}
    \caption{A deeper look into the effect of various algorithmic changes on adversarial robustness.}
    \label{fig:deep}
\end{figure}

We take a closer look into this finding in \Cref{fig:deep}. We see that in the data-limited regime, i.e., without synthetic data, the choice of the loss function (PGD vs. TRADES) or regularization (EMA vs No EMA) plays an important role in improving the adversarial robustness of a given model -- evidenced by high variance in autoattack accuracy. 
However, in the presence of synthetic data, the variance due to these choices is reduced, and model size plays a greater role. These findings are reasonable and also evident from popular leaderboards such as RobustBench~\citep{croce2020robustbench}.

\subsection{What is the Computational Burden of Adversarial Robustness?}
\label{sec:law}

\begin{figure*}[t!]
    \centering
    \begin{subfigure}{0.3\textwidth}
        \includegraphics[width=\textwidth]{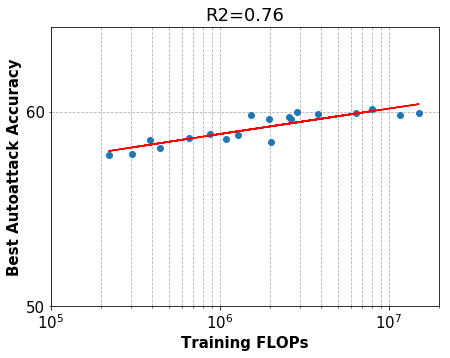}
        \caption{Best autoattack accuracy follows a power law with a small growth rate.}
        \label{fig:auto}
    \end{subfigure}
    \begin{subfigure}{0.32\textwidth}
        \includegraphics[width=\textwidth]{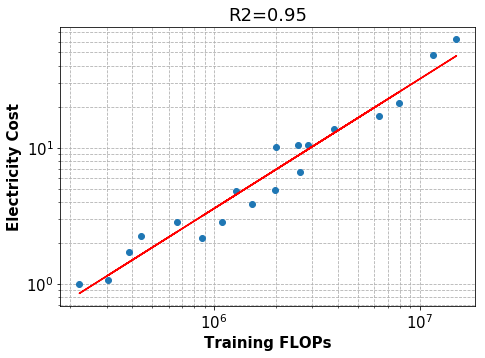}
        \caption{Electricity cost follows a power law with a high growth rate.}
        \label{fig:electricity}
    \end{subfigure}
    \begin{subfigure}{0.32\textwidth}
        \includegraphics[width=\textwidth]{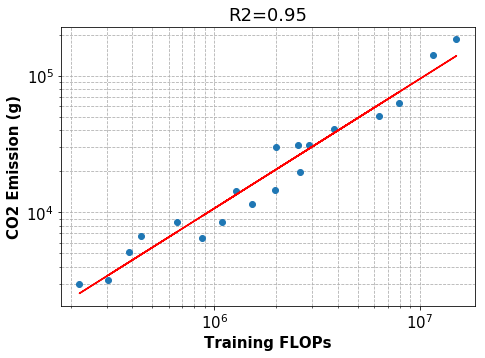}
        \caption{CO2 emission follows a power law with a high growth rate.}
        \label{fig:co2}
    \end{subfigure}
    \caption{Limits of scaling compute on adversarial robustness by analyzing best autoattack accuracy configurations in log-log space.}
    \label{fig:comp}
\end{figure*}

Having identified promising strategies for improving adversarial robustness in \Cref{sec:algo}, in this section, we explore the limits of scaling compute (i.e., FLOPs) to achieve adversarial robustness. Specifically, we study empirical scaling laws for adversarial robustness performance in terms of AutoAttack accuracy. We first divide the FLOP range into $19$ equisized bins. Then, for each FLOP bin, we determine which run achieves the best AutoAttack accuracy. This serves as the best performance envelope for our FLOP range in consideration. 
Similarly to \citet{thompson2020computational}, we expect a polynomial relationship between compute and AutoAttack accuracy: 

\begin{equation}
    \text{accuracy} = C\times (\text{FLOPs})^\alpha,
\end{equation}

where $C$ and $\alpha$ are learnable parameters. Next, we fit power laws
to estimate the best AutoAttack accuracy for any given amount of compute. 

\Cref{fig:auto} shows the growth in the AutoAttack accuracy as a function of the computation used in these models. Given that this is plotted on a log-log scale, a straight line indicates a polynomial growth in computing per unit of performance. In other words, there is a clear power law relationship between the adversarial robustness and compute spanning two orders of magnitude (with $R2=0.76$). We find that the power law relationship also holds for electricity cost (see \Cref{fig:electricity}) and CO2 emission (see \Cref{fig:co2}) both with $R2=0.95$.

\begin{tcolorbox}[colframe=black,colback=lightgray!30,boxrule=0.8pt,boxsep=4pt,left=1pt,right=1pt,top=0pt,bottom=0pt]
\noindent \emph{AutoAttack accuracy, electricity cost, and CO2 emission of adversarial training follow a power law with respect to compute (i.e., FLOPs).}
\end{tcolorbox}

Unfortunately, the growth rate for AutoAttack accuracy is very small as seen from the estimated parameters $\alpha^* = 0.01$ and $C^* = 50.86$. On the other hand, the growth rate for CO2 emission (electricity cost) is quite large, i.e., $\alpha^* = 0.95$ with $C^* = 7.0 \cdot 10^{-6}$. These findings suggest that scaling compute alone may not be able to help solve the adversarial robustness problem.

\begin{tcolorbox}[colframe=black,colback=lightgray!30,boxrule=0.8pt,boxsep=4pt,left=1pt,right=1pt,top=0pt,bottom=0pt]
\noindent \emph{Slow growth rate (w.r.t. compute) for adversarial robustness but high growth rates for CO2 emission and electricity cost imply that scaling compute is neither effective nor efficient.}
\end{tcolorbox}

Conversely, it has been observed that, for standard training, much higher growth rates hold hence suggesting that scaling compute is much more effective than it is for adversarial training.

Finally, we note, from \Cref{fig:deep}, that \emph{the usage of extra data (despite being synthetic) has significantly more advantageous scaling laws}, compared to the models trained without extra data. 

\subsection{Are Adversarial Training Techniques Brittle?}

Careful readers may have noted, at this point, that the best model among those in \Cref{fig:algo1} has only little more than $60\%$ accuracy (to be precise, $60.14\%$). They may also have noted that, among the setups we test, there is also the training setup used by~\citet{gowal2021improving}, i.e., WideResNet-70-16 with GELU activation function, 1M synthetic samples, TRADES loss with 10 steps attack, 400 epochs, and EMA. This model has a reported $63.58\%$ AutoAttack accuracy on the original paper, while our model, trained with the same set-up, has $59.92\%$ robust accuracy, more than $3.6\%$ less than the results reported in the paper.

We implemented the adversarial training pipeline from scratch, starting from the \texttt{timm} library\footnote{\url{https://github.com/huggingface/pytorch-image-models}} and by adding components based on the descriptions in the papers we reproduce, as well as by observing the code released with the papers. We also followed the suggestions made by~\citet{rade2021pytorch} (e.g., not updating mean and variance of the BatchNorm layers statistics when computing the adversarial example) to match as much as possible the original set-up by~\citep{gowal2021improving}. We double-checked our implementation and setups, including code, architectures, and hyper-parameters carefully, without any luck in exactly matching the results from~\citet{gowal2021improving}.

While we admit that exactly reproducing those results may not be impossible, it turned out to be very challenging. 
The fact that it is not possible to reproduce these results suggests that these methods may be relying on relatively brittle features, that are not really robust to small changes in the training setup. Hence, we believe that one factor to keep in account when trying to improve adversarial robustness before scaling up compute is to make sure to rely on setups that are robust to small changes in the training hyperparameters.

\section{Discussion and Limitations}

Our work presents several limitations, including the fact that we do not match exactly the results from previous work, we are only testing on a low-resolution and small-sized dataset, and we do not have confidence intervals. However, none of these limitations is easy to address in practice: we spent non-negligible time (and compute resources) to reproduce the results from the literature, and scaling up to larger datasets such as ImageNet and doing multiple runs per sample would be impractical in terms of compute (training WideResNet-76-10 for 400 epochs with 10-steps TRADES takes more than 16 days on a node with four Nvidia V100 GPUs).

Regardless, we believe that our work still gives a clear overview that scaling up compute does not give as much advantage for adversarial training as it does for standard training, especially without the usage of extra data. Thus, future research should explore the generation and the usage of extra data for improving adversarial training. We advocate for more innovative and comprehensive approaches for continued progress in improving adversarial robustness, encouraging the field to explore new directions. We firmly believe that devising new methods to achieve robust and resilient AI systems is not only necessary but also holds promise for a future where machine learning technologies can withstand adversarial threats.

\section*{Acknowledgment}
We thank Brian Bartoldson, James Diffenderfer, and Florian Tramèr for helpful comments. Edoardo Debenedetti is supported by armasuisse Science and Technology. Zishen Wan is supported in part by CoCoSys, a Semiconductor Research Corporation (SRC) JUMP 2.0 program sponsored by DARPA.
This work was performed under the auspices of the U.S. Department of Energy by the Lawrence Livermore National Laboratory under Contract No. DE- AC52-07NA27344 and was supported by the LLNL-LDRD Program under Project No. 23-ERD-030 and 22-DR-009 (LLNL-CONF-856533). 


\bibliography{refs}
\bibliographystyle{plainnat}

\end{document}